\documentclass[11pt]{article}

\usepackage[utf8]{inputenc}
\usepackage[T1]{fontenc}
\usepackage{amsmath,amssymb,amsfonts}
\usepackage{graphicx}
\usepackage{booktabs}
\usepackage{natbib}
\usepackage{hyperref}
\usepackage{cleveref}
\usepackage{microtype}
\usepackage{float}
\usepackage{subcaption}
\usepackage[section]{placeins}
\usepackage[margin=1in]{geometry}

\title{Deriving Decoder-Free Sparse Autoencoders from First Principles}
\author{Alan Oursland}
\date{}

\begin{document}

\maketitle

\begin{abstract}
Gradient descent on log-sum-exp (LSE) objectives performs implicit expectation--maximization (EM): the gradient with respect to each component output equals its responsibility. The same theory predicts collapse without volume control analogous to the log-determinant in Gaussian mixture models. We instantiate the theory in a single-layer encoder with an LSE objective and InfoMax regularization for volume control. Experiments confirm the theory's predictions. The gradient--responsibility identity holds exactly; LSE alone collapses; variance prevents dead components; decorrelation prevents redundancy. The model exhibits EM-like optimization dynamics in which lower loss does not correspond to better features and adaptive optimizers offer no advantage. The resulting decoder-free model learns interpretable mixture components, confirming that implicit EM theory can prescribe architectures.
\end{abstract}

\section{Introduction}
\label{sec:introduction}

Deep learning models are typically designed through intuition and experimentation, with theory following to explain what works. This paper inverts that order. We begin from a theoretical result, implicit expectation--maximization (EM) in neural networks, derive a model from its requirements, and test whether the prescription succeeds. The theory makes specific predictions: what architecture is required, what objective to use, and what failure modes will occur without each component. We build exactly what the theory specifies and test each prediction. The resulting model works, not because it was optimized for benchmarks, but because principled derivation yields a coherent design.

\subsection{Implicit EM Theory}
\label{sec:implicit-em-theory}

Prior work shows that gradient descent on log-sum-exp (LSE) loss results in implicit EM during backpropagation \citep{oursland2025implicit}. This occurs because the gradient with respect to each component output equals its responsibility, the posterior probability that the component explains the input. This identity is exact. The forward pass computes responsibilities via a softmax; the backward pass returns those same quantities as gradients; parameter updates therefore play the role of the M-step.

Prior work interprets neural network outputs as distances from learned prototypes, where lower values indicate better matches \citep{oursland2024mahalanobis}. Under this view, softmax is an assignment mechanism. Responsibilities sum to one across competing components, distributing each input among them. \Cref{sec:theory} develops this in detail.

The theory also exposes a limitation. Classical mixture models include a log-determinant term that prevents components from collapsing to points. Neural LSE objectives typically lack an analogous constraint. When implicit EM is present without corresponding safeguards, representations degenerate. We test this prediction directly.

\subsection{The Question}
\label{sec:question}

Theories can explain or they can prescribe. An explanatory theory redescribes existing models in new terms. This aids understanding but not necessarily building. A prescriptive theory specifies what to build from first principles and risks falsification if that specification fails.

This paper investigates if implicit EM theory can be prescriptive. The question is not whether it can be used to improve sparse autoencoders, which would treat the theory as a source of heuristics layered onto existing designs. The question is whether the theory itself specifies a working model. If we build exactly what implicit EM requires, do we obtain a functioning system? Do the predicted failure modes appear when required components are removed, and do the predicted behaviors emerge when the full system is trained?

A positive answer would establish implicit EM as a foundation for principled model design. A negative answer would confine it to post-hoc interpretation.

\subsection{This Paper}
\label{sec:this-paper}

We construct a model using implicit EM theory. The architecture is minimal, consisting of a single linear layer followed by ReLU. The LSE loss provides EM dynamics. InfoMax regularization provides volume control. Variance penalties prevent collapse. Decorrelation penalties prevent redundancy. These are neural equivalents of the log-determinant term in Gaussian mixture models. We add nothing for empirical convenience and omit nothing the theory requires.

\begin{equation}
\label{eq:objective-intro}
L = L_{\text{LSE}} + L_{\text{var}} + L_{\text{decorr}}
\end{equation}

The result is a decoder-free sparse autoencoder, or equivalently, a neural mixture model. There is no reconstruction loss, no L1 sparsity penalty, and no decoder. Sparse, interpretable features emerge through competition alone: components specialize because the objective rewards specialization.

Experiments test the theory's predictions directly. We verify the gradient--responsibility identity exactly. We show that LSE alone collapses, that variance prevents dead components, and that decorrelation prevents redundancy. The learned features are mixture components, digit prototypes rather than dictionary elements. Training dynamics exhibit EM-like properties. SGD is learning-rate insensitive and adaptive optimizers offer no advantage.

\subsection{Contribution}
\label{sec:contribution}

This paper makes four contributions.

\paragraph{Implicit EM theory is generative.} The theory specifies requirements that constrain model design. We derived an architecture and objective from those requirements. The resulting unsupervised model works, and its behavior confirms theoretical predictions.

\paragraph{Each theoretical prediction is confirmed.} The gradient--responsibility identity holds to floating-point precision. LSE alone collapses as predicted. Variance prevents dead components; decorrelation prevents redundancy. The learned features are mixture components, digit prototypes rather than unstructured projections.

\paragraph{Training dynamics are consistent with EM structure.} SGD is insensitive to learning rate across three orders of magnitude. Lower loss does not correspond to better features. Adaptive optimizers offer no advantage. These observations were not predicted in advance but are consistent with implicit EM producing a well-conditioned optimization landscape.

\paragraph{The derived model performs well.} In our experiments, the theory-derived model achieves 93.4\% probe accuracy versus 90.3\% for a standard sparse autoencoder, with half the parameters and no decoder. Limited trials prevent strong claims, but that a model built purely from theory compares favorably to heuristic designs is itself evidence that the derivation is sound.

\subsection{Roadmap}
\label{sec:roadmap}

\Cref{sec:theory} develops the theoretical foundation: distance-based representations, the LSE identity, and volume control via InfoMax. \Cref{sec:model} instantiates this prescription as a concrete model. \Cref{sec:experiments} validates the theory's predictions experimentally. \Cref{sec:discussion} discusses the implications, including the role of decoders and what the optimization results reveal.

\section{What Implicit EM Theory Requires}
\label{sec:theory}

This section develops the theoretical foundation from which our model is derived. The derivation proceeds in three steps.

\begin{enumerate}
    \item We adopt the distance-based interpretation of neural network outputs, which provides the geometric substrate for what follows (\Cref{sec:distance-representations}).
    \item We present the log-sum-exp identity. The gradient with respect to each component output equals its responsibility, allowing gradient descent to perform implicit EM (\Cref{sec:lse-identity}).
    \item Neural LSE objectives lack the volume control that prevents collapse in classical mixture models. We address this with InfoMax regularization, using variance and decorrelation penalties as the neural analogue of the log-determinant (\Cref{sec:volume-control,sec:solution}).
\end{enumerate}

The section concludes with a summary mapping each theoretical requirement to its implementation (\Cref{sec:theory-summary}).

The result is a complete specification. The theory requires distances, an LSE objective, and explicit volume control. \Cref{sec:model} instantiates this specification as a concrete model.

\subsection{Distance-Based Representations}
\label{sec:distance-representations}

The standard interpretation of neural networks treats outputs as confidences or scores where a high output indicates strong evidence for a hypothesis. This interpretation, while intuitive, obscures the geometric structure underlying what neural networks compute.

An alternative interpretation reframes outputs as distances from learned prototype regions \citep{oursland2024mahalanobis}. With ReLU, each node's prototype surface is the half-space where its output is zero. Under this view, a linear layer followed by an activation computes a deviation from a learned reference. The pre-activation $z_j = w_j^\top x + b_j$ measures how far the input lies from the decision boundary in a scaled Euclidean space.

The connection to Gaussian components is exact. The Mahalanobis distance along a principal direction $v$ with eigenvalue $\lambda$ from a mean $\mu$ is
\begin{equation}
d = |\lambda^{-1/2} v^\top (x - \mu)|
\end{equation}
which has the form $|Wx + b|$ for appropriate $W$ and $b$. The weight vector encodes both the principal direction and the precision along it. Standard ReLU networks compute this via the identity $|z| = \text{ReLU}(z) + \text{ReLU}(-z)$, decomposing signed distance into two half-space detectors.

Our model uses a single linear layer followed by ReLU, yielding the half-space distance structure described above.

The key shift is semantic. A neural network performs the same operations whether you view its output as a confidence or a distance. Probabilities are derived quantities, arising only after exponentiation and normalization transform distances into relative likelihoods.

Throughout this paper, we adopt the distance-based interpretation: \textbf{lower distance means better match}. This convention, called energy in energy-based models \citep{lecun2006tutorial}, is essential for the results that follow.

\subsection{The LSE Identity}
\label{sec:lse-identity}

Consider an encoder that maps an input $x$ to $K$ component distances $d_1(x), \ldots, d_K(x)$. Following \Cref{sec:distance-representations}, lower distance indicates a better match. Given these distances, define the log-sum-exp marginal objective:
\begin{equation}
\label{eq:lse}
L_{\text{LSE}}(x) = -\log \sum_{j=1}^{K} \exp(-d_j(x))
\end{equation}

This objective has a straightforward interpretation: it is minimized when at least one component assigns low distance to the input. It encodes the requirement that \emph{some} component must explain each data point. This is the same intuition underlying mixture models \citep{bishop2006pattern}.

The key property of this objective is an exact algebraic identity \citep{oursland2025implicit}. Taking the gradient with respect to any component distance gives:
\begin{equation}
\label{eq:gradient-responsibility}
\frac{\partial L_{\text{LSE}}}{\partial d_j}
= \frac{\exp(-d_j)}{\sum_{k=1}^{K} \exp(-d_k)}
= r_j
\end{equation}
where $r_j$ is the softmax responsibility, the posterior probability that component $j$ explains the input under the current distances.

In classical expectation--maximization \citep{dempster1977maximum}, the E-step computes responsibilities and the M-step updates parameters using responsibility-weighted statistics. These steps alternate explicitly. With an LSE objective, this separation dissolves:
\begin{itemize}
    \item The forward pass computes distances, which determine responsibilities implicitly through the softmax in \Cref{eq:gradient-responsibility}.
    \item The backward pass computes gradients, which by \Cref{eq:gradient-responsibility} are exactly those responsibilities.
    \item The optimizer step updates parameters, with each component receiving gradient signal proportional to its responsibility for the input.
\end{itemize}

Gradient descent on an LSE objective \emph{is} EM, performed continuously rather than in discrete alternating steps.

\subsection{The Volume Control Requirement}
\label{sec:volume-control}

The LSE identity (\Cref{eq:gradient-responsibility}) provides a mechanism for learning. Responsibility-weighted gradient updates implement implicit EM. But this mechanism does not specify what should be learned.

In supervised learning, labels complete the objective. Cross-entropy has LSE structure, and target labels define the desired outcome. Responsibilities are trained to match those labels.

In unsupervised learning, there are no labels. The LSE loss alone admits degenerate solutions.

\paragraph{The collapse problem.} Without volume control, the LSE loss admits degenerate solutions. One failure mode is distance collapse: the network maps all inputs to nearby points, trivializing responsibilities. Another is winner dominance: a component that captures slightly more mass receives stronger gradients, captures more mass still, and dominates entirely. This positive feedback is inherent to EM dynamics and is classically controlled by the log-determinant term.

The result in either case is a collapsed representation. Components either produce uniform output or a single component claims everything while others die. The encoder may satisfy the LSE loss, but the representation carries little information.

\paragraph{The mixture model precedent.} This failure mode is not unique to neural networks. Gaussian mixture models face the same problem, and their solution is instructive. The log-likelihood of a data point under a GMM component includes a log-determinant term:
\[
\log P(x \mid k) \propto
-\frac{1}{2}(x - \mu_k)^\top \Sigma_k^{-1}(x - \mu_k)
-\frac{1}{2}\log\det(\Sigma_k)
\]

The first term is the Mahalanobis distance. The second term, the log-determinant, penalizes components with small covariance \citep{bishop2006pattern}. Without it, a component can shrink to a point, placing arbitrarily high density on any data point it occupies. The log-determinant enforces volume. Each component must occupy a larger region of support.

\paragraph{The missing term.} Neural LSE objectives have no equivalent to the log-determinant. Without explicit volume control, nothing prevents collapse. \Cref{sec:solution} identifies the required correction.

\subsection{The Solution}
\label{sec:solution}

The log-determinant in Gaussian mixture models plays two distinct roles, each addressing a different failure mode.

\paragraph{The diagonal role.} For a covariance matrix with eigenvalues $\lambda_1, \ldots, \lambda_K$, the log-determinant is $\sum_j \log \lambda_j$. With uncorrelated components, this reduces to $\sum_j \log \text{Var}(A_j)$. As any variance approaches zero, this term diverges to negative infinity. A component cannot produce constant output without incurring unbounded penalty. The diagonal enforces existence. Every component must maintain non-zero variance.

\paragraph{The off-diagonal role.} The log-determinant also depends on correlations between components. If two components are perfectly correlated, the covariance matrix becomes singular and its determinant vanishes. More generally, correlations reduce the determinant below what uncorrelated components would achieve. The off-diagonal structure enforces diversity. Components cannot become redundant copies of one another.

InfoMax regularization follows the same structural pattern with variance and decorrelation penalties.

\paragraph{Variance penalty.} The variance penalty discourages low-variance components:
\[
L_{\text{var}} = -\sum_{j=1}^{K} \log \text{Var}(A_j)
\]

Under Gaussian assumptions, log-variance is proportional to differential entropy \citep{linsker1988self}. Maximizing this term encourages each component to carry information rather than remain constant. The logarithmic barrier ensures that collapse is forbidden. As $\text{Var}(A_j) \to 0$, the penalty diverges. This is exactly the diagonal of the log-determinant in the uncorrelated case.

\paragraph{Decorrelation penalty.} The decorrelation penalty discourages redundant components:
\[
L_{\text{tc}} = \|\text{Corr}(A) - I\|_F^2
\]

Decorrelation approximates statistical independence at the level of second-order statistics \citep{bell1995information}. The penalty is zero when components are uncorrelated and increases with off-diagonal correlations. Components that respond identically across inputs are penalized. This corresponds to the off-diagonal structure of the log-determinant.

\paragraph{Role equivalence.} The correspondence between GMM volume control and InfoMax regularization is summarized in \Cref{tab:role-equivalence}.

\begin{table}[t]
\centering
\caption{Correspondence between GMM volume control and InfoMax regularization.}
\label{tab:role-equivalence}
\begin{tabular}{lll}
\toprule
GMM term & Neural equivalent & Function \\
\midrule
$\log \det(\Sigma)$ (diagonal) & $\sum_j \log \text{Var}(A_j)$ & Prevent dead components \\
$\log \det(\Sigma)$ (off-diagonal) & $-\|\text{Corr}(A) - I\|_F^2$ & Prevent redundant components \\
\bottomrule
\end{tabular}
\end{table}

Implicit EM theory demands distance computation, an LSE loss, and explicit volume control. The first two were established in \Cref{sec:distance-representations,sec:lse-identity}. The variance and decorrelation penalties supply the third. \Cref{sec:model} assembles these elements into a concrete model.

\subsection{Summary}
\label{sec:theory-summary}

Each component of the model follows from a specific theoretical source, as shown in \Cref{tab:theory-components}.

\begin{table}[t]
\centering
\caption{Theoretical sources and implementations for each model component.}
\label{tab:theory-components}
\begin{tabular}{lll}
\toprule
Component & Theory Source & Implementation \\
\midrule
Distances & \citet{oursland2024mahalanobis} & Linear layer ($z = Wx + b$) \\
Activation & Distance interpretation & ReLU \\
EM structure & \citet{oursland2025implicit} & LSE loss (\Cref{eq:lse}) \\
Volume control (diagonal) & Log-determinant structure & Variance penalty \\
Volume control (off-diagonal) & Log-determinant structure & Decorrelation penalty \\
\bottomrule
\end{tabular}
\end{table}

The variance penalty prevents collapse. The decorrelation penalty prevents redundancy. We added nothing for empirical convenience. \Cref{sec:model} presents the full instantiation.

\section{The Derived Model}
\label{sec:model}

We first present the architecture (\Cref{sec:architecture}), then define the complete objective (\Cref{sec:objective}). We analyze the dynamics that emerge from the objective (\Cref{sec:dynamics}) and state the theoretical predictions that \Cref{sec:experiments} tests (\Cref{sec:predictions}).

\subsection{Architecture}
\label{sec:architecture}

The identity in \Cref{eq:gradient-responsibility} holds for any differentiable parameterization of the distances $d_j(x)$ \citep{oursland2025implicit}. We therefore adopt the simplest instantiation: a single linear layer followed by a nonlinearity.
\begin{equation}
\label{eq:architecture}
z = Wx + b, \qquad d = \phi(z)
\end{equation}

Here $W \in \mathbb{R}^{K \times D}$ maps inputs to $K$ components. 

We use ReLU in our experiments. Other activations can be substituted without affecting the implicit EM property. ReLU produces non-negative distances, with zero indicating a strong match.

Responsibilities need not be computed explicitly during training. By \Cref{eq:gradient-responsibility}, they appear in the gradients. For analysis, they can be recovered as
\begin{equation}
\label{eq:responsibilities}
r = \text{softmax}(-d)
\end{equation}

This is unsupervised learning. We do not reconstruct the input. There is no decoder.

\subsection{Complete Objective}
\label{sec:objective}

The objective from \Cref{eq:objective-intro} expands to
\begin{equation}
\label{eq:full-objective}
L = -\log \sum_{j=1}^{K} \exp(-d_j) - \lambda_{\text{var}} \sum_{j=1}^{K} \log \text{Var}(d_j) + \lambda_{\text{tc}} \|\text{Corr}(d) - I\|_F^2
\end{equation}

Variance and correlation are computed across batches during training.

\paragraph{LSE term.}
At least one component must explain each input. The LSE term attracts components toward data they explain well.

\paragraph{Variance penalty.}
Components must maintain non-zero variance. The hyperparameter $\lambda_{\text{var}}$ controls the strength of this constraint.

\paragraph{Decorrelation penalty.}
Prevents redundancy. Controlled by $\lambda_{\text{tc}}$.

For some architectures, a weight regularizer can help:
\begin{equation}
\label{eq:weight-reg}
L_{\text{wr}} = \lambda_{\text{wr}} \|W^\top W - I\|_F^2
\end{equation}

This encourages orthogonality among weight vectors, preventing multiple components from converging to the same direction. We do not use it in our experiments.

\subsection{What This Model Is}
\label{sec:what-model-is}

The architecture in \Cref{eq:architecture} and the objective in \Cref{eq:full-objective} admit three complementary descriptions, each illuminating a different aspect of the same system.

\paragraph{A decoder-free sparse autoencoder.}
Standard SAEs map inputs through an encoder to a sparse bottleneck, then reconstruct via a decoder, training on reconstruction loss plus an L1 sparsity penalty \citep{olshausen1996emergence,bricken2023monosemanticity}. Our model retains only the encoder with no explicit sparsity penalty. It learns sparse, interpretable features comparable to those of standard SAEs (\Cref{sec:experiments}).

\paragraph{A neural mixture model.}
Each row of $W$ defines a component. The distances $d_j(x)$ measure how well each component explains the input. The LSE objective is the negative log marginal likelihood. The InfoMax terms play the role of the log-determinant. This is not an analogy. The mathematics are identical; only the parameterization differs.

\paragraph{A minimal test case.}
The model exists to test the theory. \Cref{eq:architecture} is the simplest architecture that computes distances. \Cref{eq:full-objective} is the objective the theory specifies. We add no architectural embellishments, auxiliary losses, or heuristics.

\subsection{Dynamics}
\label{sec:dynamics}

The objective balances attraction against structure. This explains why the model learns useful representations rather than collapsing.

\subsubsection{The LSE Term Is Attractive}

The LSE term pulls components toward data they explain well. The gradient identity ensures this attraction is responsibility-weighted. Left unchecked, this leads to collapse (\Cref{sec:volume-control}).

\subsubsection{The InfoMax Terms Are Structural}

The variance and decorrelation penalties constrain \emph{how} components can reduce their distance, without dictating \emph{where} they should go. Variance enforces selectivity. Decorrelation enforces diversity. Together, they shape attraction into a stable equilibrium.

\subsubsection{The Equilibrium Is Competitive Coverage}

Components distribute themselves to tile the data manifold, each specializing in a region of input space. A component cannot expand its territory without reducing its variance or overlapping with others, both of which are penalized. The result is a soft partition of the input space.

This resembles competitive learning in self-organizing maps \citep{kohonen1982self}, but with soft responsibilities rather than winner-take-all assignments.

\subsubsection{Sparsity Is Emergent}

The objective contains no explicit sparsity penalty. Sparse representations nevertheless arise. As components specialize, each input activates only a few components. Sparsity emerges from structure.

\subsection{Theoretical Predictions}
\label{sec:predictions}

The theory makes five testable predictions.

\begin{table}[t]
\centering
\caption{Theoretical predictions and their sources.}
\label{tab:predictions}
\begin{tabular}{ll}
\toprule
Prediction & Theoretical Source \\
\midrule
Gradient equals responsibility exactly & LSE identity (\Cref{eq:gradient-responsibility}) \\
LSE alone collapses & Missing volume control (\Cref{sec:volume-control}) \\
Variance term prevents dead units & Diagonal of log-determinant (\Cref{sec:solution}) \\
Decorrelation prevents redundancy & Off-diagonal of log-determinant (\Cref{sec:solution}) \\
Features are mixture components & Implicit EM interpretation (\Cref{sec:lse-identity}) \\
\bottomrule
\end{tabular}
\end{table}

\paragraph{Prediction 1: Gradient equals responsibility.}
\Cref{eq:gradient-responsibility} is an algebraic identity. It should hold to numerical precision.

\paragraph{Prediction 2: LSE alone collapses.}
Without volume control, all components should die.

\paragraph{Prediction 3: Variance prevents collapse.}
Adding variance penalty should prevent dead components. Without decorrelation, components may still be redundant.

\paragraph{Prediction 4: Decorrelation prevents redundancy.}
Adding decorrelation should force components to encode distinct information.

\paragraph{Prediction 5: Features are mixture components.}
Learned features should resemble prototypes that compete for data. Visualized features should exhibit global structure.

Confirming all five would support implicit EM theory as a foundation for model design.

\section{Experimental Validation}
\label{sec:experiments}

We test each prediction in turn, then report an exploratory observation about training dynamics.

\subsection{Experiment 1: Theorem Verification}
\label{sec:exp-theorem}

\paragraph{Prediction.} The gradient equals the responsibility exactly (\Cref{eq:gradient-responsibility}).

\paragraph{Method.} We verify the identity with a single forward-backward pass. We generate random activations $a \in \mathbb{R}^{64 \times 128}$ (64 samples, 128 components), compute the LSE loss $L = -\sum_i \log \sum_j \exp(-a_{ij})$, compute responsibilities $r = \text{softmax}(-a)$, and backpropagate to obtain gradients. We then compare \texttt{a.grad} and $r$ element-wise across all 8,192 values.

No training is involved. The test isolates the mathematical identity from any learned parameters.

\begin{figure}[!htbp]
\centering
\includegraphics[width=0.6\columnwidth]{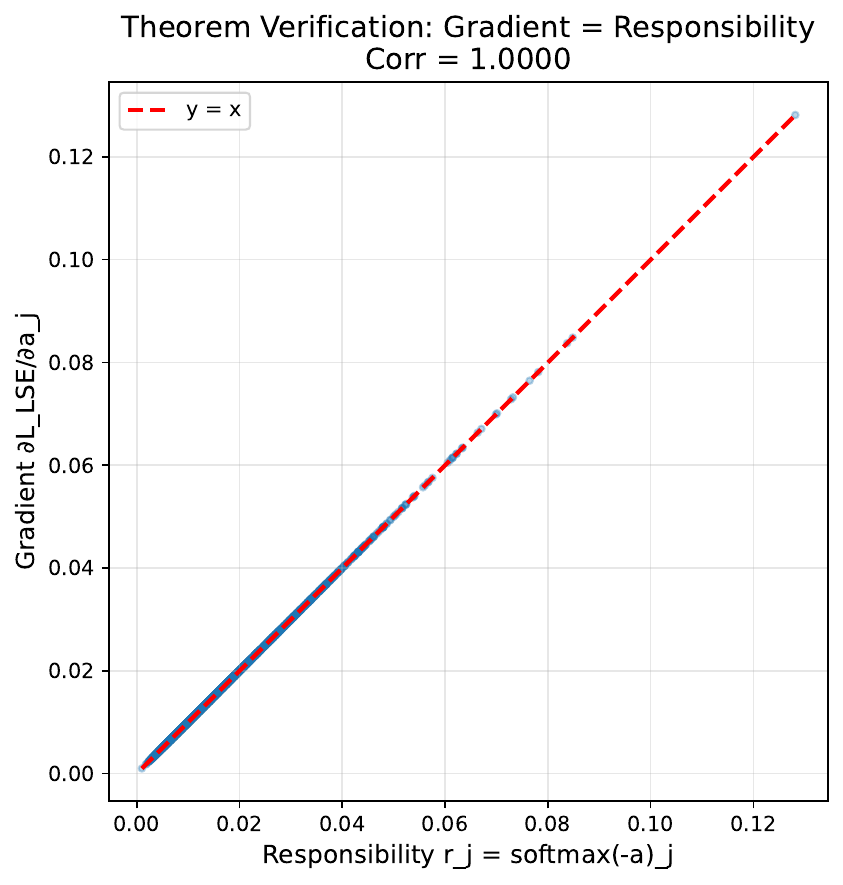}
\caption{Gradient vs.\ responsibility for 8,192 random activations. All points lie on $y = x$, confirming the identity in \Cref{eq:gradient-responsibility} to floating-point precision.}
\label{fig:theorem}
\end{figure}

\paragraph{Results.} \Cref{fig:theorem} plots the gradient $\partial L_{\text{LSE}} / \partial a_j$ against the responsibility $r_j = \text{softmax}(-a)_j$ for all 8,192 values. All points lie on the $y=x$ line. The correlation is 1.0000; the maximum absolute error is $4.47 \times 10^{-8}$; the mean absolute error is $2.79 \times 10^{-9}$. These deviations are at floating-point precision.

\paragraph{Interpretation.} Points on the y=x line indicate identical values. Prediction 1 is confirmed.

\subsection{Experiment 2: Ablation Study}
\label{sec:exp-ablation}

\paragraph{Predictions.} 
\begin{itemize}
    \item \emph{LSE only:} Complete collapse.
    \item \emph{LSE + variance:} No dead units, but redundancy.
    \item \emph{LSE + variance + decorrelation:} Stable, diverse representations.
    \item \emph{Variance + decorrelation only:} Viable representations but different dynamics.
\end{itemize}

\paragraph{Method.}
We train four configurations on MNIST with 64 components, 100 epochs, batch size 128, Adam at learning rate 0.001, and $\lambda_{\text{var}} = \lambda_{\text{tc}} = 1.0$ when enabled. Three random seeds per configuration (\Cref{tab:ablation-configs}).

\begin{table}[t]
\centering
\caption{Ablation configurations.}
\label{tab:ablation-configs}
\begin{tabular}{lccc}
\toprule
Configuration & LSE & Variance & Decorrelation \\
\midrule
LSE only & \checkmark & & \\
LSE + var & \checkmark & \checkmark & \\
LSE + var + tc & \checkmark & \checkmark & \checkmark \\
var + tc only & & \checkmark & \checkmark \\
\bottomrule
\end{tabular}
\end{table}

We measure dead units (components with variance $< 0.01$), redundancy ($\|\text{Corr}(A) - I\|_F^2$), and responsibility entropy ($\mathbb{E}_x[H(r(x))]$, where higher values indicate softer competition).

\paragraph{Results.} \Cref{tab:ablation-results} summarizes the outcomes.

\begin{table}[t]
\centering
\caption{Ablation study results.}
\label{tab:ablation-results}
\begin{tabular}{lccc}
\toprule
Configuration & Dead Units & Redundancy & Resp.\ Entropy \\
\midrule
LSE only & 64/64 (100\%) & --- & 4.16 \\
LSE + var & 0/64 (0\%) & 1875 & 3.77 \\
LSE + var + tc & 0/64 (0\%) & 29 & 3.85 \\
var + tc only & 0/64 (0\%) & 28 & 1.99 \\
\bottomrule
\end{tabular}
\end{table}

\paragraph{Interpretation.}

\emph{LSE only.} All 64 components died. The loss plateaued by epoch 10. Without volume control, collapse is complete.

\emph{LSE + variance.} No dead units, but redundancy explodes to 1875. Components are alive but nearly identical. Variance prevents collapse but not redundancy.

\emph{LSE + variance + decorrelation.} Adding decorrelation reduces redundancy by $64\times$ (from 1875 to 29). The full objective achieves zero dead units, low redundancy, and high responsibility entropy.

\emph{Variance + decorrelation only.} Without LSE, the model still achieves zero dead units and low redundancy. However, responsibility entropy drops from 3.85 to 1.99, indicating sharper competition. The LSE term provides soft responsibilities characteristic of mixture models. Without it, the model performs whitening rather than soft clustering.

Predictions 2–4 confirmed.

\subsection{Experiment 3: Benchmark Comparison}
\label{sec:exp-benchmark}

\paragraph{Goal.}
Assess whether the theory-derived model learns useful representations. We compare against a standard sparse autoencoder \citep{olshausen1996emergence,bricken2023monosemanticity}.

\paragraph{Method.}
We train two models on MNIST with matched hidden dimension, 100 epochs, batch size 128, Adam at learning rate 0.001, and five random seeds (\Cref{tab:benchmark-models}). The SAE uses L1 weight 0.01.

\begin{table}[t]
\centering
\caption{Model configurations for benchmark comparison.}
\label{tab:benchmark-models}
\begin{tabular}{llll}
\toprule
Model & Architecture & Loss & Parameters \\
\midrule
Theory-derived (ours) & Linear (784$\to$64) + ReLU & LSE + InfoMax & 50,240 \\
Standard SAE & Linear (784$\to$64) + ReLU + Linear (64$\to$784) & MSE + L1 & 101,200 \\
\bottomrule
\end{tabular}
\end{table}

We evaluate on four metrics:
\begin{itemize}
    \item \emph{Linear probe accuracy:} Freeze the encoder, train a linear classifier (sklearn LogisticRegression) on the features, report MNIST test accuracy.
    \item \emph{L0 sparsity:} Fraction of features active (nonzero) per input.
    \item \emph{Parameters:} Total trainable parameters.
    \item \emph{Reconstruction MSE:} For SAE, use the trained decoder. For our model, use $W^\top$ as a pseudo-decoder (no decoder is trained).
\end{itemize}

\paragraph{Results.} See \Cref{tab:benchmark-results}.

\begin{table}[t]
\centering
\caption{Benchmark comparison results.}
\label{tab:benchmark-results}
\begin{tabular}{lcc}
\toprule
Metric & Theory-Derived (Ours) & Standard SAE \\
\midrule
Linear Probe Accuracy & \textbf{93.43\% $\pm$ 0.38\%} & 90.26\% $\pm$ 0.32\% \\
L0 Density & \textbf{26.8\%} (17.2/64) & 50.3\% (32.2/64) \\
Parameters & \textbf{50,240} & 101,200 \\
Reconstruction MSE & 0.143 $\pm$ 0.001 & \textbf{0.026 $\pm$ 0.001} \\
\bottomrule
\end{tabular}
\end{table}

\paragraph{Interpretation.}

\emph{Feature quality.}
The theory-derived model reaches 93.4\% linear probe accuracy versus 90.3\% for the SAE. For reference, logistic regression on raw MNIST pixels achieves approximately 92\%. The learned features outperform raw pixels; the SAE features underperform them.

\emph{Sparsity.}
Despite no explicit sparsity penalty, the theory-derived model activates only 27\% of features per input versus 50\% for the SAE trained with L1. Decorrelation induces sparsity indirectly: uncorrelated components must specialize on different inputs.

\emph{Parameters.}
Eliminating the decoder halves the parameter count.

\emph{Reconstruction.}
The SAE achieves lower reconstruction error. This is expected: it optimizes reconstruction with a trained decoder, whereas we use the untrained transpose $W^\top$. The probe results show that information is preserved, but encoded for discrimination rather than pixel-wise reconstruction.

The theory-derived model compares favorably to the SAE baseline: higher probe accuracy, sparser representations, and half the parameters. Both are baseline implementations without extensive tuning. Standard techniques (learning rate schedules, data augmentation, architectural refinements) could likely improve both. The point is not that our model beats an optimized SAE, but that a model derived purely from theory performs well without benchmark-driven tuning.

\subsection{Experiment 4: Feature Visualization}
\label{sec:exp-features}

\paragraph{Prediction.}
If the model performs implicit EM, learned features should resemble mixture components, prototypes that compete for data, rather than dictionary elements that combine additively \citep{olshausen1996emergence}. Encoder weights should show global structure (whole patterns) rather than local parts (edges or strokes).

\paragraph{Method.}
We visualize encoder weights from both models by reshaping each row of $W \in \mathbb{R}^{64 \times 784}$ into a $28 \times 28$ image. A diverging colormap shows positive weights (blue), negative weights (red), and zero (white). Each model is scaled to its own maximum absolute weight.

\begin{figure}[!htbp]
\centering
\begin{subfigure}[t]{0.48\columnwidth}
\centering
\includegraphics[width=\textwidth]{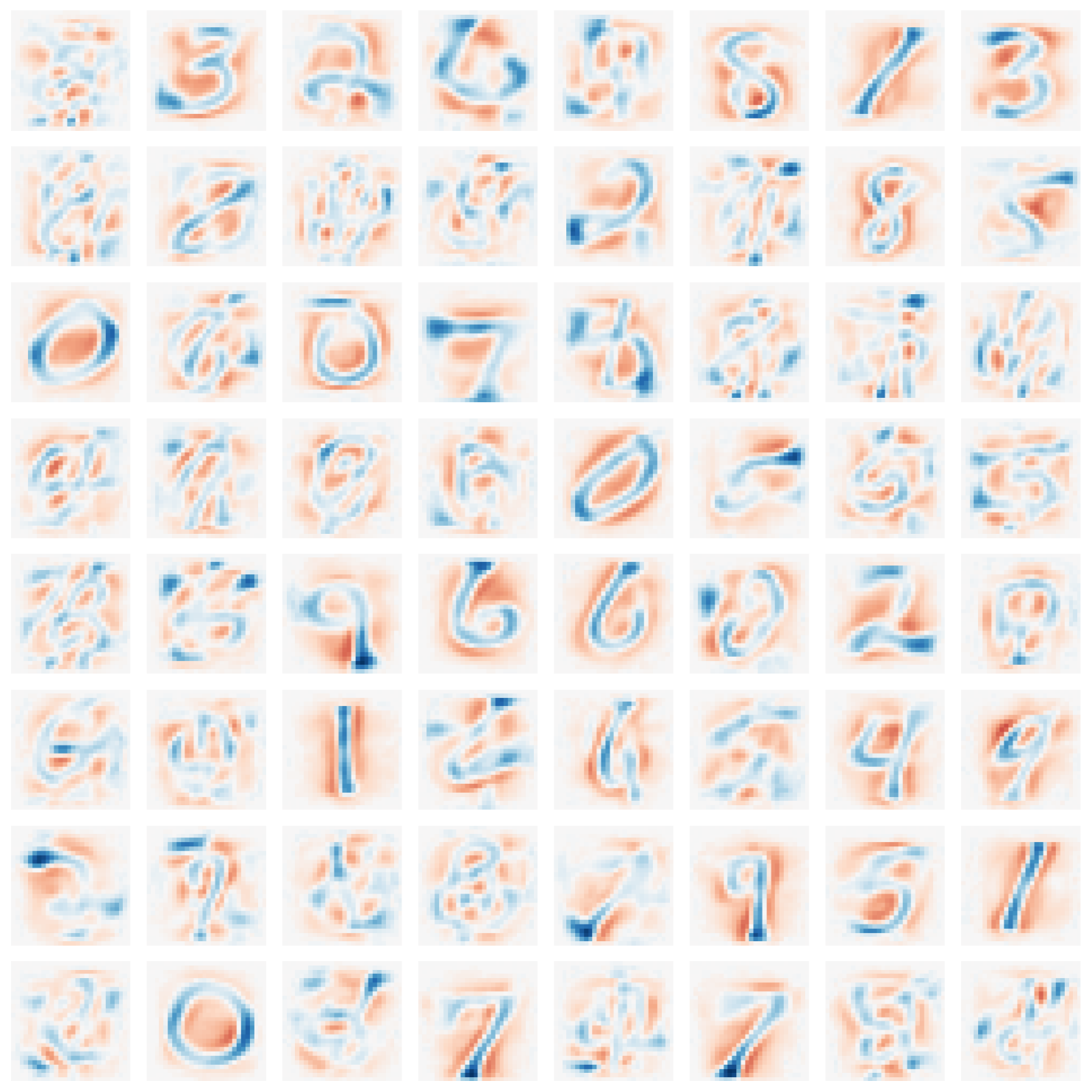}
\caption{Theory-derived model}
\label{fig:features-ours}
\end{subfigure}
\hfill
\begin{subfigure}[t]{0.48\columnwidth}
\centering
\includegraphics[width=\textwidth]{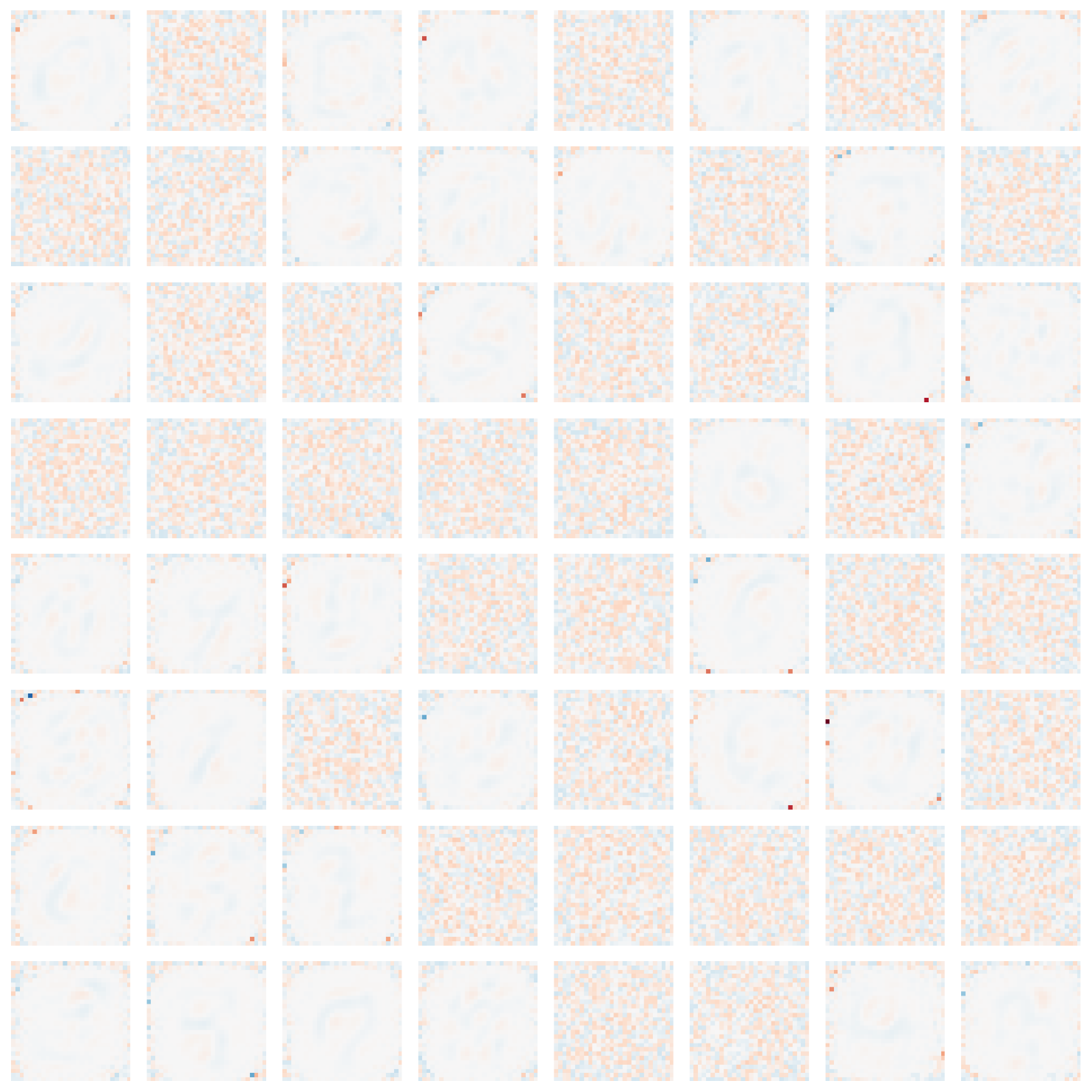}
\caption{Standard SAE}
\label{fig:features-sae}
\end{subfigure}
\caption{Learned encoder weights. (a) Theory-derived model: features form recognizable digit prototypes with center-surround structure, consistent with mixture components competing for data. (b) Standard SAE: encoder weights show little interpretable structure; representational content is carried primarily by the decoder.}
\label{fig:features}
\end{figure}

\paragraph{Results.}
\Cref{fig:features-ours,fig:features-sae} show the learned encoder weights for both models.

The theory-derived model (\Cref{fig:features-ours}) learns clear digit prototypes. Across the 64 features, multiple variants of each digit appear: circular 0s, vertical and slanted 1s, loopy and angular 2s, distinct forms of 3s through 9s. Many features exhibit center-surround structure, a digit-shaped region of one sign surrounded by the opposite sign, indicating that each component acts as both detector and suppressor. All features are active and interpretable; no dead units or degenerate patterns.

The standard SAE encoder (\Cref{fig:features-sae}) shows largely unstructured weights. Most features resemble low-magnitude noise, with only faint digit-like structure visible in a subset. The SAE encoder weights are qualitatively noisier and less organized.

\paragraph{Interpretation.}
The theory-derived model learns prototypes because it is a mixture model. Each row of $W$ defines a component; the LSE objective induces soft competition; the InfoMax terms enforce distinctness. The weights resemble GMM centroids because that is what implicit EM produces.

The SAE encoder weights lack structure because the encoder need only produce activations the decoder can invert. Structure resides in the decoder.

This explains the probe accuracy gap in \Cref{sec:exp-benchmark}. Because encoder weights are organized by digit class, a linear probe effectively reads out class identity. A probe on SAE features must compose unstructured activations into class predictions, a harder task.

\subsection{Experiment 5: Training Dynamics}
\label{sec:exp-dynamics}

\paragraph{Motivation.}
We trained all previous experiments with Adam per standard practice. Out of curiosity, we tried SGD and were surprised by the results.

\paragraph{Method.}
We train the model using SGD and Adam across learning rates spanning three orders of magnitude ($10^{-4}$ to $10^{-1}$), with three random seeds per configuration (24 runs total). Other settings match \Cref{sec:exp-ablation}. We analyze loss trajectories and evaluate feature quality via linear probe accuracy.

\begin{figure}[!htbp]
\centering
\includegraphics[width=\columnwidth]{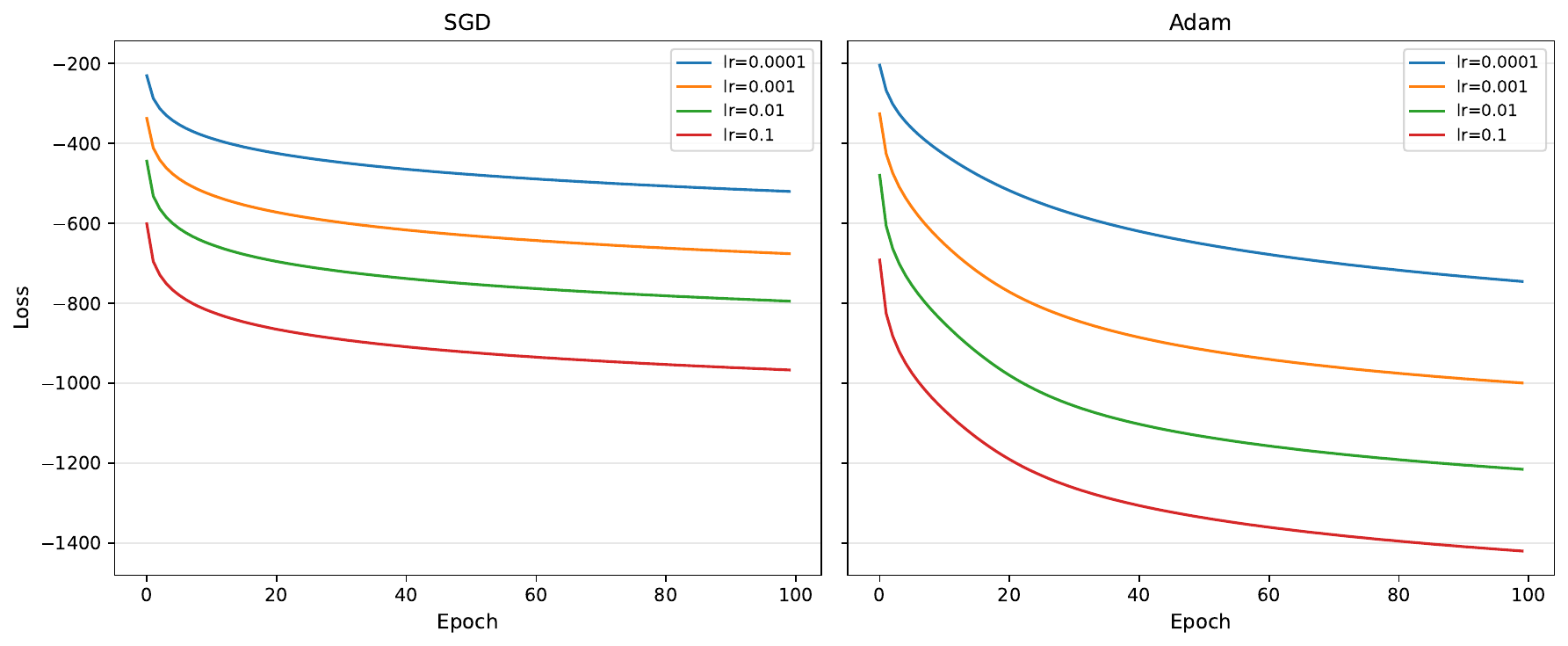}
\caption{Loss curves for SGD (left) and Adam (right) across learning rates. SGD curves plateau; Adam curves continue descending. Despite achieving substantially lower loss, Adam produces features of comparable quality.}
\label{fig:dynamics}
\end{figure}

\paragraph{Results.} See \Cref{tab:dynamics,fig:dynamics}.

\begin{table}[t]
\centering
\caption{Training dynamics results across optimizers and learning rates.}
\label{tab:dynamics}
\begin{tabular}{llcc}
\toprule
Optimizer & lr & Final Loss & Probe Acc \\
\midrule
SGD & 0.0001 & $-520 \pm 1$ & 92.2\% $\pm$ 0.3\% \\
SGD & 0.001 & $-676 \pm 0$ & 92.6\% $\pm$ 0.2\% \\
SGD & 0.01 & $-795 \pm 29$ & 93.6\% $\pm$ 0.1\% \\
SGD & 0.1 & $-967 \pm 18$ & 93.5\% $\pm$ 0.1\% \\
Adam & 0.0001 & $-745 \pm 2$ & 92.9\% $\pm$ 0.0\% \\
Adam & 0.001 & $-999 \pm 1$ & 93.5\% $\pm$ 0.4\% \\
Adam & 0.01 & $-1215 \pm 21$ & 93.5\% $\pm$ 0.2\% \\
Adam & 0.1 & $-1420 \pm 13$ & 93.1\% $\pm$ 0.2\% \\
\bottomrule
\end{tabular}
\end{table}

\paragraph{Finding 1: Adam's usual advantage disappears.}
On most objectives, Adam outperforms SGD, especially at low learning rates. That pattern does not appear here. SGD at lr = $10^{-4}$, normally impractically slow, reaches a stable solution within 100 epochs. The adaptive machinery provides little benefit.

\paragraph{Finding 2: Lower loss does not imply better features.}
Adam achieves nearly 50\% lower loss than SGD at lr = 0.1. Yet probe accuracy is indistinguishable: 93.1\% versus 93.5\%. The objective admits directions that reduce loss without affecting downstream utility.

\paragraph{Finding 3: SGD is insensitive to learning rate.}
Across a $1000\times$ range, SGD yields probe accuracy between 92.2\% and 93.6\%. Learning rate affects where SGD settles, but not whether it finds good features.

\paragraph{Interpretation.}
These results point to a well-conditioned objective. Responsibility weighting normalizes gradient magnitudes across components. This is precisely what adaptive optimizers approximate. When the objective provides it directly, Adam has little to add.

The analogy to classical EM is suggestive. Explicit EM has no learning rate because the M-step computes optimal updates given current responsibilities. Implicit EM introduces a step size, but if the gradient direction is already correct, scaling changes speed rather than outcome.

The remaining anomaly, Adam lowering loss without improving features, suggests the objective contains degrees of freedom orthogonal to representation quality. Characterizing this null space remains open.

\subsection{Summary}
\label{sec:exp-summary}

\Cref{tab:validation-summary} summarizes the results.

\begin{table}[t]
\centering
\caption{Summary of experimental validation.}
\label{tab:validation-summary}
\begin{tabular}{lll}
\toprule
Prediction & Experiment & Result \\
\midrule
Gradient = responsibility & Theorem & Exact ($10^{-8}$ error) \\
LSE alone collapses & Ablation & 100\% dead units \\
Variance prevents death & Ablation & 0\% dead units \\
Decorrelation prevents redundancy & Ablation & $64\times$ reduction \\
Features are mixture components & Visualization & Digit prototypes \\
\midrule
\textbf{Observation} & & \\
Model compares favorably to SAE & Benchmark & +3.1\% probe accuracy \\
SGD learning-rate insensitive & Dynamics & 92--94\% across $1000\times$ \\
\bottomrule
\end{tabular}
\end{table}

All predictions confirmed.

\section{Discussion}
\label{sec:discussion}

All of our EM theory predictions were confirmed through experimental results. This section considers what that means.

\subsection{What This Validates}
\label{sec:what-validates}

Implicit EM theory is generative. We built what it required. The model works, and it works for the reasons the theory predicts.

The experiments confirm every prediction (\Cref{tab:validation-summary}).

The theory came first. The experiments tested whether the model it specifies actually works. It does. And not just in the binary sense—the qualitative behavior also matches. Soft competition via responsibilities, competitive coverage, emergent sparsity without L1. These weren't fitted; they were predicted.

\subsection{Why It Works}
\label{sec:on-performance}

The theory-derived model compared favorably to the baseline. Higher probe accuracy, sparser representations, half the parameters. We expected a working model, perhaps a worse one. So why did it perform well?

Standard sparse autoencoders are assembled from compensatory mechanisms. The decoder enforces information preservation because without it, the encoder could discard information. The L1 penalty enforces sparsity because without it, all features would activate densely. The reconstruction loss anchors features to input fidelity because without it, representations could drift arbitrarily. Each component exists to counteract a failure mode that would otherwise emerge.

The result is a system at war with itself. Reconstruction wants features that combine to reproduce the input. L1 wants features that do not activate. The encoder wants to distribute information across features. The decoder wants to invert whatever the encoder produces. Training navigates these conflicting pressures, settling into a compromise that satisfies none of them fully.

Our model has no such conflicts. The LSE term attracts components toward data. The variance penalty prevents collapse. The decorrelation penalty prevents redundancy. Each term does one thing. The terms do not fight each other.

This may explain the performance gap. When the objective is internally coherent, optimization is straightforward. When the objective is a collection of patches, optimization is a negotiation. The SAE's encoder learns to produce something the decoder can invert while the L1 penalty tries to shut it down. Our encoder just learns prototypes.

A caveat: both models are baseline implementations. The SAE uses untied weights and standard L1 regularization. Many improvements exist. The theory-derived model uses the simplest architecture that satisfies the theoretical requirements. The point is not that principled derivation always beats heuristic engineering. The point is that it can, and when it does, the reason is coherence.

\subsection{Reconstruction and Volume Control}
\label{sec:reconstruction}

The decoder has been a fixture of autoencoder architectures since their introduction \citep{hinton2006reducing}. Sparse autoencoders inherited this structure, adding L1 penalties while retaining the encoder-decoder-reconstruction framework \citep{vincent2008extracting}. The decoder appeared necessary. Remove it, and representations degenerate.

Why did representations degenerate? The standard explanation is that reconstruction enforces information preservation. Without pressure to reproduce the input, the encoder could discard information arbitrarily. This explanation is correct but incomplete. It describes what the decoder does without asking whether reconstruction is the only way to achieve it.

Reconstruction implicitly supplies volume control.

\paragraph{Anti-collapse.} If a feature never activates, it contributes nothing to reconstruction. Inputs that need that feature incur higher reconstruction error, producing gradient signal that revives the dead unit. Reconstruction prevents collapse because dead features degrade reconstruction. This is the diagonal of the log-determinant: components must have non-zero variance.

\paragraph{Anti-redundancy.} If two features encode identical information, the decoder can use either one. Using both offers no benefit. The L1 penalty then suppresses the redundant copy, since it incurs sparsity cost without improving reconstruction. Reconstruction combined with L1 discourages redundancy because redundant features are wasteful. This is the off-diagonal of the log-determinant: components must be decorrelated.

The decoder provides volume control indirectly, through the reconstruction bottleneck. We provide it directly, through explicit regularization. Same function, different mechanism.

With explicit volume control, the decoder becomes unnecessary. The variance penalty prevents collapse without relying on reconstruction to revive dead units. The decorrelation penalty prevents redundancy without relying on L1 to prune duplicates. The functions the decoder served are now enforced by the objective itself.

And something interesting happens. Properties that SAEs achieve through opposing forces emerge naturally from coherent ones.

Sparsity, for instance. SAEs achieve sparsity through L1, which penalizes activation. The objective wants features to activate for reconstruction; L1 wants them to stay off. Sparsity emerges from this fight. In our model, sparsity emerges from specialization. When components carve up the input space, most inputs fall in the territory of only a few. The others simply are not relevant. Sparsity as consequence of structure.

Information preservation, similarly. SAEs verify that information survives by demanding reconstruction. If the decoder cannot reproduce the input, something was lost. Our model ensures information survives directly. Variance means each feature carries information. Decorrelation means features carry different information. No decoder needed to check.

Feature visualization makes this concrete (\Cref{sec:exp-features}). Our encoder weights are interpretable prototypes. Each row of $W$ forms a recognizable digit pattern with center-surround structure. The SAE encoder weights are largely unstructured, with only faint patterns visible under close inspection.

This asymmetry is revealing. In the SAE, the decoder carries the representational burden. The encoder merely produces signals the decoder can invert. Interpretability is not required of the encoder because structure resides in the decoder. In our model, the encoder is the representation. There is no decoder to compensate for an unstructured projection. The encoder must learn structure, and it does.

The decoder was compensatory. It patched a deficiency in the objective by supplying volume control implicitly through reconstruction. The log-determinant's functions, smuggled in through the back door. Once the deficiency is addressed directly, the patch is unnecessary. The decoder was never about reconstruction. It was about preventing collapse and redundancy. Those functions now reside where they belong.

\subsection{Connection to Self-Supervised Learning}
\label{sec:self-supervised}

The goals of our InfoMax terms appear in recent self-supervised methods, discovered independently through extensive empirical research. Barlow Twins \citep{zbontar2021barlow} and VICReg \citep{bardes2022vicreg} both prevent collapse and redundancy through explicit regularization.

VICReg makes the structure particularly clear: variance, invariance, covariance. The variance term keeps feature standard deviations above a threshold. The covariance term penalizes off-diagonal entries of the covariance matrix. Same goals as our variance and decorrelation penalties, different implementations. Where they use invariance to augmentations as the primary objective, we use LSE attraction.

This convergence from different directions is striking. The self-supervised community arrived at these constraints through years of careful experimentation on large-scale vision tasks, systematically identifying what prevents representation collapse. We arrived at them through analogy to the log-determinant in mixture models. That both paths lead to variance and decorrelation suggests this structure is fundamental.

The implicit EM framework offers a theoretical interpretation for their empirical findings: these regularizers enforce volume control. Conversely, the success of Barlow Twins and VICReg at scale provides evidence that our derivation, validated here only on MNIST, rests on solid foundations. Their work suggests the approach scales. Our work suggests why it works.

\subsection{On Optimization}
\label{sec:on-optimization}

The training dynamics experiment revealed something odd. SGD plateaus by roughly epoch 70 regardless of learning rate. Adam keeps descending, reaching nearly 50\% lower loss, but feature quality stays the same. What is going on?

Classical EM has no learning rate. The algorithm alternates between computing responsibilities and updating parameters to maximize expected log-likelihood. Step size is fixed by the mathematics. Convergence occurs when responsibilities stabilize, in a number of iterations determined by the problem.

Implicit EM may inherit this behavior. The gradient of the LSE objective equals the responsibility. SGD follows this gradient directly. When responsibilities stabilize, gradients vanish and SGD stops. Learning rate scales how fast you get there, but not when you arrive. This would explain why SGD at $10^{-4}$ and SGD at $10^{-1}$ reach similar feature quality: they trace the same path at different speeds.

Adam complicates the picture. Its adaptive per-parameter scaling is designed for ill-conditioned objectives. But if responsibility weighting already normalizes gradients across components, the landscape is well-conditioned by construction. Adam's machinery becomes unnecessary. Worse, near equilibrium its second-moment estimate shrinks as gradients become small, effectively increasing step size when it should decrease. The optimizer keeps pushing when it should stop.

The puzzle is what Adam is optimizing. Loss keeps dropping. Features do not improve. This suggests the objective has degrees of freedom orthogonal to representation quality. Directions in parameter space that reduce loss without changing what the model learns. What are they? We do not know. Characterizing this null space remains open.

A caveat: this is speculation. The observations are consistent with implicit EM producing a well-conditioned landscape where simple optimizers suffice. Responsibility structure and learning-rate invariance co-occur; whether one causes the other remains open. The connection to classical EM is suggestive.

\subsection{Limitations}
\label{sec:limitations}

This work validates implicit EM theory on a minimal test case.

\paragraph{Scale.} The model is a single linear layer with 64 components, trained on MNIST. The theory holds for any differentiable parameterization of distances, but we have not tested deeper architectures, larger datasets, or hundreds of components. The competitive dynamics that work here may behave differently at scale.

\paragraph{LLM activations untested.} A primary motivation for sparse autoencoders is interpretability of large language models \citep{bricken2023monosemanticity,cunningham2023sparse}. We tested on images. The features learned on MNIST are digit prototypes, visually interpretable. Whether similar structure emerges for linguistic or abstract features is an open question.

\paragraph{Formal EM connection incomplete.} We call this implicit EM based on the gradient-responsibility identity. But we have not derived closed-form M-step updates, proved convergence guarantees, or established the precise relationship between gradient descent and EM fixed-point iterations. The interpretation remains empirical.

\paragraph{Hyperparameters not tuned.} We used $\lambda_{\text{var}} = \lambda_{\text{tc}} = 1.0$ throughout with no systematic search. The SAE baseline used L1 weight 0.01 without tuning. Different settings might narrow or widen the performance gap. Our goal was to validate the theory.

We tested the theory in one setting. Its scope may be broader. Deeper networks, larger scale, language model activations, and formal EM connections remain for future work.

\subsection{Future Directions}
\label{sec:future}

This theory suggests several paths to explore.

\paragraph{Explicit EM.} Can we derive closed-form M-steps for the linear case? Gradient descent works, but explicit EM might converge faster, require no learning rate, and come with guarantees.

\paragraph{Supervised regularization.} Cross-entropy already has LSE structure. Adding InfoMax during supervised learning would encourage mixture-like representations even with labels. Structured features alongside discriminative ones.

\paragraph{Layer-wise pretraining.} Hinton and others showed this works \citep{hinton2006reducing,bengio2006greedy}. The field moved on when end-to-end backprop proved sufficient. But reconstruction was always an awkward objective for intermediate layers. LSE+InfoMax is principled. Each layer learns a mixture model over its inputs. Hierarchical mixtures, all the way up.

\paragraph{Conditioning pretrained models.} Apply LSE+InfoMax to representations learned by supervised training. How does the unsupervised structure differ? Fine-tune and measure how far supervision pushes from the mixture solution.

\paragraph{Activation geometry.} ReLU gives half-space prototypes, with zero distance for an entire region. Softplus gives smooth, always-positive distance. What happens to mixture component character when the geometry changes? Do sharp boundaries produce sharper prototypes? Does smoothness help or hurt specialization? A systematic study across activations could reveal whether the choice matters more than we assumed.

\paragraph{Attention.} Prior work established that attention is implicit EM—softmax over key similarities produces responsibilities, and value vectors receive responsibility-weighted updates \citep{oursland2025implicit}. The natural question is whether InfoMax regularization transfers. Decorrelation across attention heads could address the known redundancy problem where heads learn similar patterns. Variance penalties could prevent head collapse. The same volume control that stabilizes mixture components might stabilize attention.

\paragraph{Anomaly detection.} High LSE loss means no component explains the input well. This is a natural anomaly score, available without modification. Does it outperform reconstruction error? Does it catch different failure modes? The framework hands us an anomaly detector for free—someone should test whether it works.

\paragraph{Optimal transport.} Responsibilities distribute inputs across components. Optimal transport distributes mass across sinks. The LSE objective may be minimizing a transport cost we have not identified. If so, OT convergence guarantees might transfer, and Sinkhorn-like algorithms might accelerate training.

\paragraph{Interpretability.} Apply the framework post-hoc to trained models. Compute responsibilities for existing representations. Do pretrained vision models already have mixture structure hiding in their activations? If so, the framework becomes a lens for understanding what networks have already learned.

We have validated the theory in one setting. These directions suggest it may reach further.

\section{Conclusion}
\label{sec:conclusion}

Implicit EM theory specifies a model: compute distances, optimize a log-sum-exp objective, and include volume control. We built exactly what the theory requires, a single-layer encoder with an LSE objective and InfoMax regularization, and tested every prediction the framework makes.

All predictions were confirmed. The gradient--responsibility identity holds to floating-point precision. LSE alone collapses exactly as predicted; variance prevents dead components; decorrelation prevents redundancy. The learned features are digit prototypes, mixture components competing for data, rather than the unstructured encoder projections produced by standard sparse autoencoders. The theory-derived model reaches 93.4\% probe accuracy with half the parameters, comparing favorably to heuristic designs that rely on decoders and L1 penalties.

The optimization behavior is distinctive. Adam offers no advantage; lower loss does not produce better features; SGD is insensitive to learning rate across three orders of magnitude. These results suggest that implicit EM structure yields an unusually well-conditioned objective, with degrees of freedom that decouple loss minimization from representational quality.

The broader implication is methodological. Implicit EM theory is generative. It specifies what to build. We derived an architecture and objective directly from theory, implemented them without modification, and obtained a model that works for the reasons the theory predicts. This establishes implicit EM as a viable foundation for principled model design.

\bibliographystyle{plainnat}
\bibliography{references}

\end{document}